\documentclass[letterpaper, 10 pt, conference]{ieeeconf}  % Comment this line out if you need a4paper

\pdfoutput=1
\IEEEoverridecommandlockouts                              % This command is only needed if 
\usepackage[linesnumbered,ruled]{algorithm2e}
\usepackage{float}
\usepackage{graphics} % for pdf, bitmapped graphics files
\usepackage{epsfig} % for postscript graphics files
\usepackage{mathrsfs}
\usepackage{times} % assumes new font selection scheme installed
\usepackage{amsmath} % assumes amsmath package installed
\usepackage{amssymb}  % assumes amsmath package installed

\usepackage{amssymb}% http://ctan.org/pkg/amssymb
\usepackage{pifont}% http://ctan.org/pkg/pifont
\usepackage{graphicx}
\usepackage{pifont}
\usepackage{booktabs}

\newcommand{\NoOne}[1]{\textcolor{red}{#1}}
\newcommand{\NoTwo}[1]{\textcolor{green}{#1}}
\newcommand{\NoThree}[1]{\textcolor{blue}{#1}}

\newcounter{RNum}
\renewcommand{\theRNum}{\arabic{RNum}}
\newcommand{\Remark}{\noindent\textit{\textbf{Remark}~\refstepcounter{RNum}\textbf{\theRNum}: }}
\usepackage{color}
\usepackage{color}
\usepackage[table]{xcolor}
\usepackage[T1]{fontenc}
\usepackage{bm}
\usepackage{cite}
\usepackage{hyperref}
\usepackage{float}
\usepackage[caption=false,font=footnotesize]{subfig}
\usepackage{multirow}
\hypersetup{
	colorlinks=true,
	linkcolor=blue,
	filecolor=blue,      
	urlcolor=blue,
	citecolor=cyan,
}

\title{\LARGE \bf
SiamAPN++: Siamese Attentional Aggregation Network for Real-Time UAV Tracking} 
\author{Ziang Cao$^{1}$, Changhong Fu$^{2,*}$, Junjie Ye$^{2}$, Bowen Li$^{2}$, and Yiming Li$^{3}$% <-this % stops a space
	\thanks{$^{*}$Corresponding Author}
	\thanks{$^{1}$Ziang Cao is with the School of Automotive Studies, Tongji University, 201804 Shanghai, China.}%
	\thanks{$^{2}$Changhong Fu, Junjie Ye, and Bowen Li are with the School of Mechanical Engineering, Tongji University, 201804 Shanghai, China.
		{\tt\small changhongfu@tongji.edu.cn}}%
	\thanks{$^{3}$Yiming Li is with the Tandon School of Engineering, New York University, NY 11201 New York, United States.}%}
}

\begin{document}

\maketitle
\thispagestyle{empty}
\pagestyle{empty}

%%%%%%%%%%%%%%%%%%%%%%%%%%%%%%%%%%%%%%%%%%%%%%%%%%%%%%%%%%%%%%%%
%%%%%%%%%%%%%%%%%%%%% Section 0: abstract %%%%%%%%%%%%%%%%%%%%%%
%%%%%%%%%%%%%%%%%%%%%%%%%%%%%%%%%%%%%%%%%%%%%%%%%%%%%%%%%%%%%%%%Siamese
\begin{abstract}
Recently, the Siamese-based method has stood out from multitudinous tracking methods owing to its state-of-the-art (SOTA) performance. Nevertheless, due to various special challenges in UAV tracking, \textit{e.g.}, severe occlusion and fast motion, most existing Siamese-based trackers hardly combine superior performance with high efficiency. To this concern, in this paper, a novel attentional Siamese tracker (SiamAPN++) is proposed for real-time UAV tracking. By virtue of the attention mechanism, we conduct a special attentional aggregation network (AAN) consisting of self-AAN and cross-AAN for raising the representation ability of features eventually. The former AAN aggregates and models the self-semantic interdependencies of the single feature map via spatial and channel dimensions. The latter aims to aggregate the cross-interdependencies of two different semantic features including the location information of anchors. In addition, the anchor proposal network based on dual features is proposed to raise its robustness of tracking objects with various scales. Experiments on two well-known authoritative benchmarks are conducted, where SiamAPN++ outperforms its baseline SiamAPN and other SOTA trackers. Besides, real-world tests onboard a typical embedded platform demonstrate that SiamAPN++ achieves promising tracking results with real-time speed.

\end{abstract}
%%%%%%%%%%%%%%%%%%%%%%%%%%%%%%%%%%%%%%%%%%%%%%%%%%%%%%%%%%%%%%%%
%%%%%%%%%%%%%%%%%%%%% Section 1: INTRODUCTION %%%%%%%%%%%%%%%%%%
%%%%%%%%%%%%%%%%%%%%%%%%%%%%%%%%%%%%%%%%%%%%%%%%%%%%%%%%%%%%%%%%
\section{Introduction}
Visual object tracking is a fundamental and challenging task, whose purpose is to track the indicated object frame by frame. By virtue of the powerful flexibility of unmanned aerial vehicles (UAVs), UAV tracking has drawn considerable attention in many fields such as aerial cinematography~\cite{8968163}, path planning~\cite{8967602}, and self-localization~\cite{9457090}.
Despite great efforts, designing an efficient, accurate, and robust tracker for UAV remains an extremely challenging task. On the one hand, the limited computational resource on the embedded platform hardly meets the requirement of existing robust but computation-consuming methods. On the other hand, the UAV tracking also suffers from various special challenges brought by the mobile platform, \textit{e.g.}, fast motion, low resolution, and severe occlusion.

%
%The objective of aerial tracking is to predict the location of the object in the following frames based on its initial state. One remarkable difference between aerial tracking and general tracking is that aerial tracking requires real-time speed and low energy consumption due to the resource-constrained aerial platforms. Besides, aerial tracking suffers from various challenging scenarios introduced by unmanned aerial vehicle (UAV), \textit{e.g.}, fast motion, low resolution, and severe occlusion. Based on the aforementioned property of aerial tracking, one question is raised naturally: \textit{can we find a good balance between efficiency and robustness, and develop an efficient and effective aerial tracker?} 
Generally, there are two mainstream methods in the field of UAV tracking, \textit{i.e.}, correlation filter (CF)-based method and deep learning (DL)-based method. According to~\cite{fu2020correlation}, online CF-based trackers are widely adopted in UAV tracking due to their low computational complexity~\cite{huang2019learning, Li_2020_CVPR}. In spite of high efficiency, the accuracy and robustness of CF-based trackers fail to meet the requirement of practical UAV tracking in complex scenes. Meantime, albeit the DL-based methods have also achieved significant advancement in tracking performance, the large computation cost limits their practical application. Consequently, it is urgent to develop an efficient DL-based structure to balance performance and speed.

%In literature, aerial tracking approaches are mainly divided into two types: correlation filter-based online trackers which are CPU friendly~\cite{huang2019learning, Li_2020_CVPR, LifanICRA2020}, and deep learning-based offline trained trackers that need a high-end GPU~\cite{LiICRA2020, Fu2019IROS}. Though the former is low-cost and energy-efficient, the performance still has a clear gap compared to the latter taking advantage of the offline training. However, the latter is suffering from low efficiency. To achieve a satisfactory balance between performance and speed, this work tries to improve deep trackers' performance while lowering their redundancy.
\begin{figure}[t]
	\centering
	% 调整比例，添加图片的相对位置
	
	\includegraphics[scale=0.38]{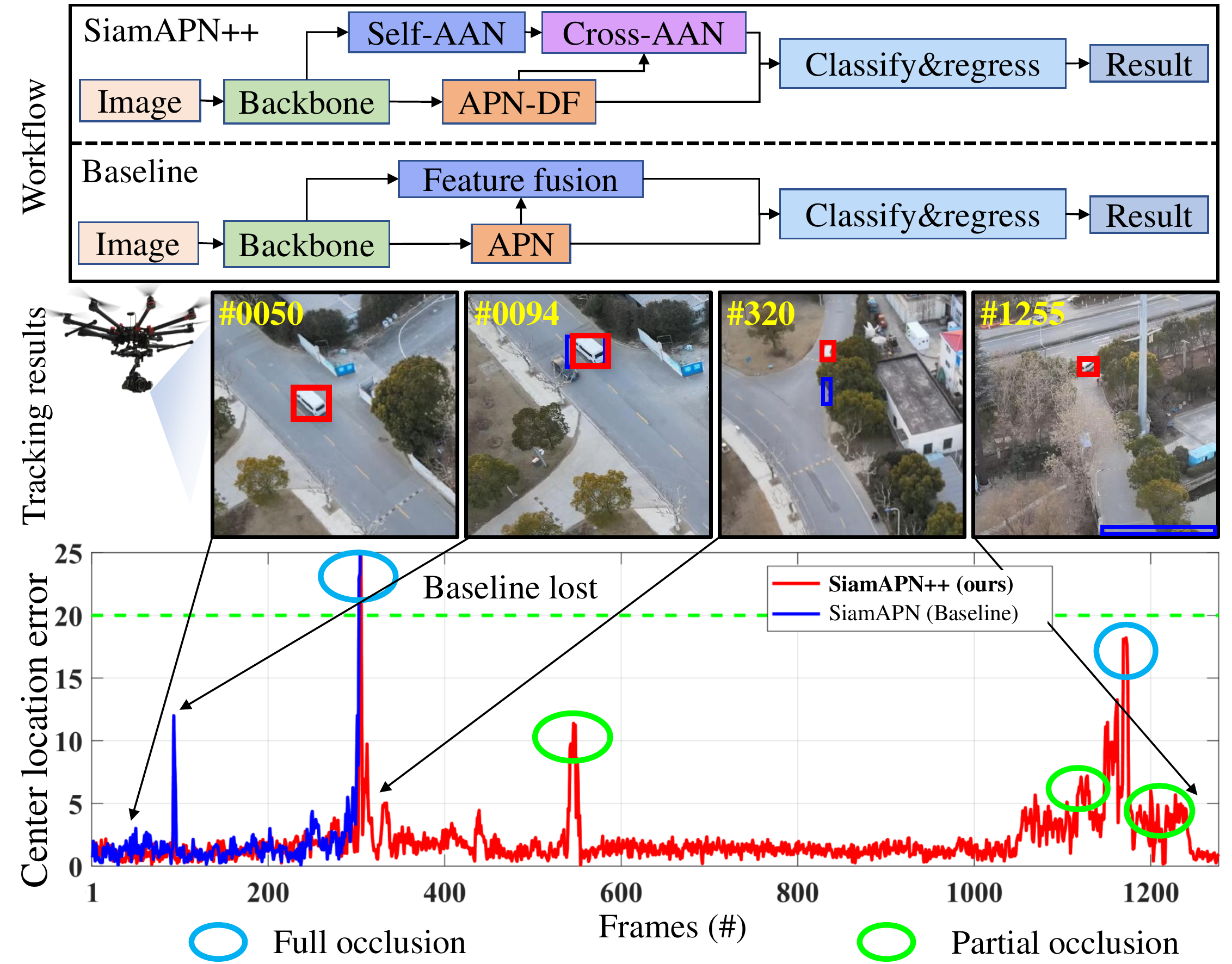}

	\caption{Comparison between the baseline tracker and the proposed tracker, \textit{i.e.}, \textcolor[rgb]{0  0  1}{SiamAPN} and \textcolor[rgb]{1  0  0}{SiamAPN++}. The figures from the top to bottom are workflow, tracking results, and center location error comparison. When facing the constantly occluded object, the attentional aggregation network (AAN) can emphasize and aggregate the effective information selectively, avoiding interference of environment.}

%		anchors-based method (\textcolor[rgb]{1 0.75 0}{SiamRPN++}~\cite{8954116}) and anchor-free method (\textcolor[rgb]{0  0  1}{SiamFC++}~\cite{xu2020siamfc++}) on $car7$ from UAV123@10fps~\cite{Mueller2016ECCV}. The \textcolor[rgb]{0  1  0}{green} box represents the ground truth bounding box. The adaptive anchors generated by APN can improve the perception of \textcolor[rgb]{1  0  0}{SiamAPN}, making our tracker straightforward to notice the occluded car. Meanwhile, the APN merely generates one no-prior anchor at each point on feature map, reducing a large number of anchors and hyper-parameters significantly.
	
    \vspace{-2pt}
	\label{fig:1} % 写 \label 跟在 \caption 后面，之后使用 \ref{}引用 enjoy their superiority due to
\end{figure}

Among the DL-based trackers, the Siamese-based trackers enjoy their superiority due to their high potential in object tracking. The Siamese network structure is widely spread by SiamFC~\cite{bertinettofully}. It proposed a new tracking strategy, \textit{i.e.}, tracking by computing the similarity between template and search patches. Then, the anchor-based method is proposed in SiamRPN~\cite{Li_2018_CVPR}. By introducing the region proposal network (RPN), it can obtain accurate bounding boxes. Based on RPN, further researches have been made to improve the tracking performance~\cite{8954116,zhu2018distractor,9156275}. Since those anchors are pre-defined, the performance of anchor-based trackers is seriously influenced by the hyper-parameters associated with anchors. To improve the generalization of trackers, the anchor-free method is proposed, which generally predicts the offset between ground truth and center point~\cite{xu2020siamfc++,9157720,cao2021hift}. Although the anchor-free method avoids the hyper-parameters, the phenomenon of imbalanced samples is still not properly solved. To solve this problem, a new anchor-free method is raised by the anchor proposal network (APN)~\cite{fu2020siamese}. However, it is not good enough to handle semantic information variation.

Meantime, the attention mechanism has also drawn much attraction in object tracking~\cite{8578608,9156275}. Unfortunately, the relationship and interdependencies of different features are generally neglected, impeding the robustness under various conditions. Therefore, we propose a novel attentional Siamese tracker namely SiamAPN++ for real-time UAV tracking, mainly consisting of the attentional aggregation network (AAN) and anchor proposal network based on dual features (APN-DF) illustrated in Fig.~\ref{fig:1}. The AAN is divided into two subnetworks, \textit{i.e.}, self-AAN and cross-AAN. The former aims to model the self-semantic interdependencies in a single feature map via spatial and channel dimensions while the latter focuses on emphasizing the interdependent channels of different feature maps adaptively. Besides, to promote the robustness of tracking objects with various scales, dual features are introduced in APN. By exploiting the interdependencies of features from different levels, APN-DF can generate more appropriate anchors than before. Real-world tests onboard the embedded platform shown in Fig.~\ref{fig:1} strongly prove the superior accuracy and robustness of SiamAPN++ while keeping a comparable speed to SiamAPN.

The main contributions of this work can be summarized as follows:

\begin{itemize}
	\item  A novel AAN is introduced to aggregate the self-semantic interdependencies of the single feature map via two mechanisms and the cross-interdependencies from different feature maps adaptively.
	
	\item The dual features are adopted in APN-DF, thereby improving the anti-interference and robustness of the proposed anchors when facing severe scale variation.

	\item Extensive evaluations on two challenging UAV tracking benchmarks prove the superior performance of SiamAPN++ especially in fast motion, low resolution, and severe occlusion. In addition, real-world tests conducted onboard an embedded platform strongly demonstrate impressive practicability and performance of our tracker with real-time speed.

\end{itemize}
%%%%%%%%%%%%%%%%%%%%%%%%%%%%%%%%%%%%%%%%%%%%%%%%%%%%%%%%%%%%%%%%
%%%%%%%%%%%%%%%%%%%%% Section 2: RELATED WORK %%%%%%%%%%%%%%%%%%
%%%%%%%%%%%%%%%%%%%%%%%%%%%%%%%%%%%%%%%%%%%%%%%%%%%%%%%%%%%%%%%%
\section{Related Works}\label{sec:RELATEDWORK}
%\subsection{Real-time tracking for UAV}
Previously, CF-based trackers have attracted much attention since MOSSE~\cite{5539960}. Depending on the high efficiency and expansibility, CF-based trackers can be deployed directly on UAVs. There is no doubt that the CF-based approaches promoted the development of UAV tracking with satisfying speed~\cite{Fu_2020_TGRS,Li_2020_CVPR}. However, the online tracking strategy hinders inevitably the structure of trackers, hardly maintaining satisfying tracking performance in practical conditions.

The Siamese-based network also shows its huge potential in the domain of object tracking. SINT~\cite{tao2016sint} firstly view the tracking task as matching the patch problem. Since the appearance of SiamFC~\cite{bertinettofully}, the advantage of the Siamese network has been obvious. It aims to measure the similarity between template and search patches by employing a fully convolutional neural network. Then, SiamRPN~\cite{8579033} introduced the RPN into the Siamese framework, dividing the tracking task into classification and regression. DaSiamRPN~\cite{zhu2018distractor} proposes a novel training method, further improving the tracking performance. Moreover, SiamRPN++~\cite{8954116} makes it possible to utilize deeper networks as backbone. Despite obtaining state-of-the-art (SOTA) performance, those anchor-based trackers above suffer from hyper-parameters and imbalanced samples. In order to eliminate these problems, anchor-free trackers are proposed, \textit{e.g.}, SiamFC++~\cite{xu2020siamfc++} and SiamCAR~\cite{9157720}. By redesigning the regression, the generalization of the trackers is raised. However, the influence caused by imbalanced samples is still existing. SiamAPN~\cite{9477413} brings a novel approach to handle the two problems at the same time, \textit{i.e.}, proposing adaptive anchors. It increases the proportion of positive samples while adopting the no-prior structure of APN.

In recent, much attention has been paid to the attention mechanism in many fields. For capturing long-range dependencies, the non-local block is proposed in~\cite{8578911}. Then, DANet~\cite{8953974} develops the self-attention mechanism to address the scene segmentation task. In visual tracking, RASNet~\cite{8578608} adopts three attention branches to adapt the model without updating the model online in visual tracking. SiamAttn~\cite{9156275} also exploits the attention mechanism module for providing an implicit manner to update the template. Besides, ECA-Net~\cite{9156697} proposes an efficient channel attention method for the deep convolutional neural network. However, those methods mentioned above merely focus on self-attention or the relationship between template and searches, neglecting the cross-interdependence between different level features.

In this work, the cross-AAN is proposed for exploring the potential of the cross-semantic interdependencies contained in different level features. Plus the self-interdependencies of the single feature map, the AAN can raise the representation ability of features effectively for handling the special challenges in UAV tracking. Besides, the APN based on dual feature structure is reconstructed for raising the anti-interference and robustness of proposing anchors and provide the reliable internal feature for AAN.

\begin{figure*}[t]
	\centering
	% 调整比例，添加图片的相对位置
	%\includegraphics[scale=0.5]{images/1.pdf}
	\scalebox{1}[1]{\includegraphics[width=1\textwidth]{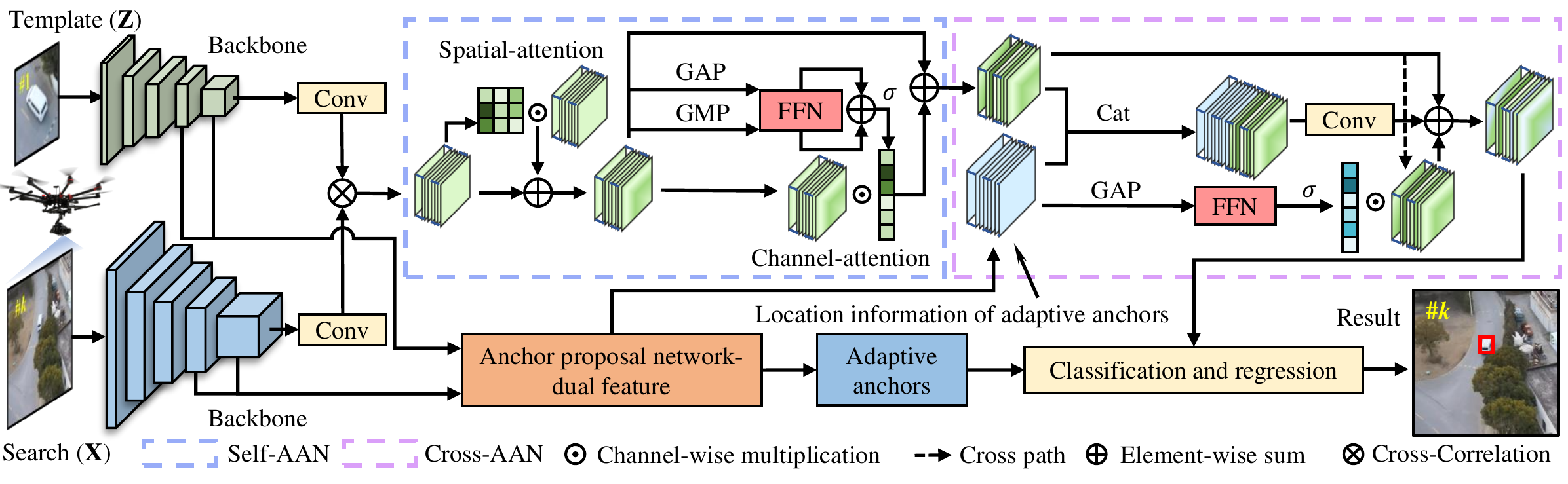}}
	%\scalebox{0.5}[0.5]{\includegraphics[trim={0 75 0 55},clip]{images/1.pdf}}
	% 写标题
	\caption{The overview of the SiamAPN++ tracker. It composes of four subnetworks, \textit{i.e.}, feature extraction network (backbone), classification and regression network, anchor proposal network-dual feature (APN-DF), and attentional aggregation network (AAN). (Best viewed in color version)
  	}
    \vspace{-10pt}
	\label{fig:main} % 写 \label 跟在 \caption 后面，之后使用 \ref{}引用
\end{figure*}

%%%%%%%%%%%%%%%%%%%%%%%%%%%%%%%%%%%%%%%%%%%%%%%%%%%%%%%%%%%%%%%%
%%%%%%%%%%%%%%%%%%%%% Section 3: Proposed method %%%%%%%%%%%%%%%%%%
%%%%%%%%%%%%%%%%%%%%%%%%%%%%%%%%%%%%%%%%%%%%%%%%%%%%%%%%%%%%%%%%
%证明featurefusion的必要性   消融实验 证明在visual trackingdataset训练uav数据集测试必要性
\section{Methodology}\label{sec:me}
\subsection{Revisit SiamAPN}\label{sec:baseline}
In this section, our baseline is revisited in brief, consisting of four subnetworks, \textit{i.e.}, feature extraction network, APN, feature fusion network, and classification\&regression network. Different from pre-defined anchors, the APN adopts a single feature map to generate adaptive anchors, avoiding the hyper-parameters associated with those pre-defined anchors. Besides, it also decreases the number of negative samples, alleviating the phenomenon of imbalanced samples. 

Although SiamAPN has achieved competitive performance, it remains two shortcomings: \textit{a}) the performance is influenced easily by complex semantic information variation. \textit{b}) the original APN hardly maintains satisfying robustness when facing objects with various scales.

%容易被深层信息误导，对多尺度的物体鲁棒性不好
%注意力          多重特征

\subsection{Proposed method}\label{sec:proposed}
In this section, the proposed SiamAPN++ is introduced in detail. As shown in Fig. \ref{fig:main}, SiamAPN++ consists of four subnetworks, \textit{i.e.}, the feature extraction network, APN-DF, AAN, and classification\&regression network. 
\subsubsection{Feature extraction network}
For fulfilling the practical deployment requirement, AlexNet~\cite{krizhevsky2012imagenet} is chosen as the backbone of the SiamAPN++ tracker. In our tracker, the feature maps from the last two layers are utilized in the tracking task. For clarification, the template image, search image, and the output of the $k^{th}$ layer are denoted as $\mathbf{Z}$, $\mathbf{X}$, and $\varphi_k(\cdot)$ respectively.

\subsubsection{APN-DF}
For the sake of boosting the robustness of the proposed anchors, the dual features from the feature extraction network are adopted to handling objects with different scales. Generally, the high-level features are concentrate on semantic features which are good for classifying while the low-level ones focus on detailed features which are helpful for accurately distinguishing objects. Therefore, we design a new APN structure to discover the internal relationship between the two different feature maps. In this way, we combine the dual features while avoiding the extra computation brought by separately calculating for each feature map.

Specifically, the $\varphi_4(\mathbf{X})$ and $\varphi_4(\mathbf{Z})$ is used to produce the similarity map as:
\begin{equation}
\mathbf{R}_{4}=\mathcal{F}(\varphi_4(\mathbf{X})\star\varphi_4(\mathbf{Z}))~,
\end{equation}
where the $\star$ represents the depth-wise cross-correlation layer introduced in~\cite{8954116} and $\mathcal{F}$ denotes the convolutional operation for decreasing the number of channels. Additionally, the similarity of the fifth layer is acquired as:
\begin{equation}
\mathbf{R}_{5}=\mathcal{F}(\varphi_5(\mathbf{X}))\star\mathcal{F}(\varphi_5(\mathbf{Z}))~.
\end{equation}

Then, the feature maps $\mathbf{R}_{\rm A}\in\mathbb{R}^{C\times H \times W}$ can be formulated as:
\begin{equation}
\begin{aligned}
\mathbf{R}_{\rm A}=\mathbf{R}_{5}+&\gamma_{1}*\mathrm{FFN}(\mathrm{GAP}(\mathbf{R}_{4}))*\mathbf{R}_{5}
\\+&\gamma_{2}*\mathcal{F}(\mathrm{Cat}(\mathbf{R}_{4},\mathbf{R}_{5}))~,
\end{aligned}
\end{equation}
where $\gamma_{k},~k=\{1,2\}$, GAP, and Cat represent the learning weights, global average pooling, and channel-wise concatenation respectively. After obtaining the $\mathbf{R}_{\rm A}$, the anchors can be obtained through two convolutional operations.

\Remark Since the dual feature structure exploits and integrates the preponderance of different level features, the robustness and anti-interference ability under scale variation are raised. Besides, it can also provide comprehensive position information for AAN.

%
% to generate anchors
%
%Different from pre-defined anchors, the location and size of proposed anchors are flexible to adapt the variable scenes during tracking, especially in low resolution, fast motion, and occlusion. Moreover, the semantic information of adaptive anchors also contributes to accurate classification and robust regression. 

\subsubsection{AAN}
AAN consists of self-AAN and cross-AAN, designing for effectively improving the representation ability of feature maps via adaptively enhancing self-semantic interdependencies and aggregating the cross-interdependencies. 

\noindent\textbf{Self-AAN:}
The self-AAN aims to enhance the self-semantic information of the single feature map via spatial and channel attention. The spatial attention is inspired by~\cite{8953974}. To explore different semantic information, we use different convolution layer to calculate $\mathbf{R'}_{5}$ as follow:
\begin{equation}
\mathbf{R'}_{5}=\mathcal{F}(\varphi_5(\mathbf{X}))\star\mathcal{F}(\varphi_5(\mathbf{Z}))~.
\end{equation}

Then, by operating three different convolution layers, three new feature maps can be generated denoted as $\{\mathbf{R}^{\rm q},\mathbf{R}^{\rm k},\mathbf{R}^{\rm v}\}\in\mathbb{R}^{C\times H \times W}$. By reshaping $\mathbf{R}^{\rm q}$ and $\mathbf{R}^{\rm k}$ to $\mathbb{R}^{C\times (H \times W)}$, cooperating matrix multiplication, and softmax layer, the spatial attention map $\mathbf{R}^{\rm a}\in\mathbb{R}^{(H \times W)\times (H \times W)}$ can be calculated. Therefore, the output of spatial attention $\mathbf{R}^{\rm s}\in\mathbb{R}^{C\times H \times W}$ can be obtained as:
\begin{equation}
\mathbf{R}^{\rm s}=\gamma_{3}\mathbf{R}^{\rm v}\times \mathbf{R}^{\rm a}+\mathbf{R'}_{5}~,
\end{equation}
where $\gamma_{3}$ is a learning weight and $\times$ denotes the matrix multiplication.

For exploiting the interdependencies between channels and improve the representation ability of features, we also build the channel-attention branch. To obtain exhaustive channel enhancement, global max-pooling (GMP) and GAP are adopted before FFN. Besides, the channel attention FFN is share-weight to explore the interdependencies among channels more effectively. Therefore, the calculation process of $\mathbf{R}_{\rm c}\in\mathbb{R}^{C\times H \times W}$ is as follows:

\begin{equation}
\begin{aligned}
&\mathbf{w}=\mathrm{FFN}(\mathrm{GAP}(\mathbf{R}^{\rm s}))+\mathrm{FFN}(\mathrm{GMP}(\mathbf{R}^{\rm s}))\\
&\mathbf{R}_{\rm c}=\mathbf{R}^{\rm s}+\gamma_{4}\mathrm{Sigmoid}(\mathbf{w})*\mathbf{R}^{\rm s}~,
\end{aligned}
\end{equation} 
where $\gamma_{4}$ is a learning weight and $*$ denotes the channel-wise multiplication.

\Remark By building the spatial and channel attention, the self-AAN can aggregate and model the self-semantic interdependencies of the single feature map. Therefore, it can provide stable and robust self-attentional features for cross-AAN.

\noindent\textbf{Cross-AAN:}
Since the anchors are generated adaptively, semantic information about anchors is necessary for SiamAPN++ to locate the anchors before predicting. Therefore, the location information of anchors is quite significant for classification and regression performance. To aggregate the cross-interdependencies between two different features and integrate the location information of anchors, cross-AAN is introduced.

There are two cross paths in cross-AAN. One path applies a feedforward neural network (FFN) structure to generate a corresponding weight for each channel of $\mathbf{R}_{\rm c}$, aggregating the interdependencies among channels explicitly. Besides, before the FFN, GAP is cooperated to compress the feature to vector. The other path aims to emphasize the interdependent channels of $\mathbf{R}_{\rm A}$ and $\mathbf{R}_{\rm c}$ implicitly via element-wise concatenation. Therefore, the final feature maps $\mathbf{R}\in\mathbb{R}^{C\times H \times W}$ can be computed as:
\begin{equation}
\begin{aligned}
\mathbf{R}=\mathbf{R}_{\rm c}+&\gamma_{5}*\mathrm{FFN}(\mathrm{GAP}(\mathbf{R}_{\rm A}))*\mathbf{R}_{\rm c}
\\+&\gamma_{6}*\mathcal{F}(\mathrm{Cat}(\mathbf{R}_{\rm A},\mathbf{R}_{\rm c}))~,
\end{aligned}
\end{equation}
where $\gamma_{k},~k=\{5,6\}$ are also learning weights to adaptively maintain the balance between the different branches.
\begin{figure}[t]
	\centering
	\includegraphics[width=1\linewidth]{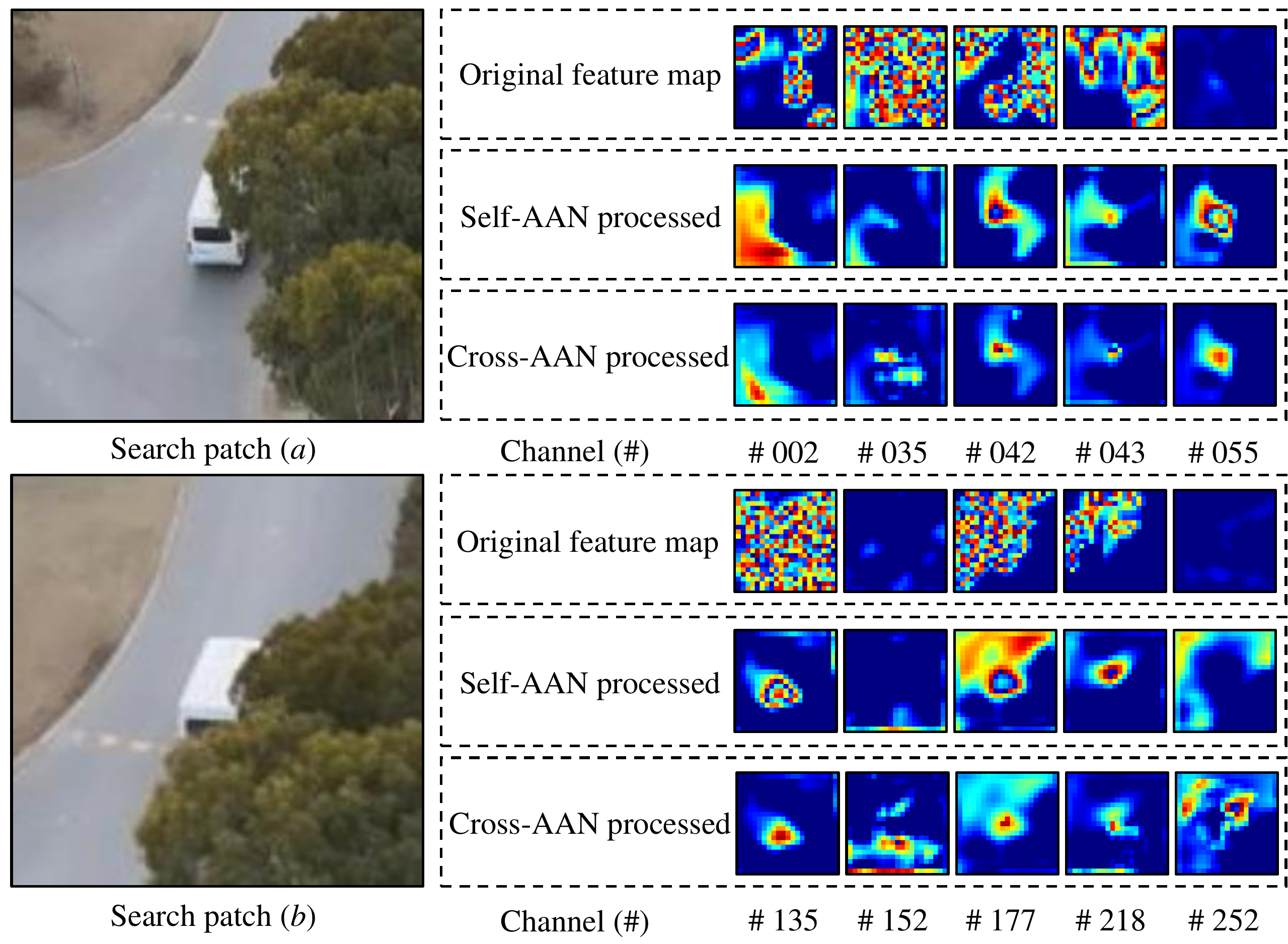}
	\caption{Visualization of the AAN output feature maps in the real-world test. It shows that AAN indeed provides effective and robust feature maps for tracking occluded objects accurately.}
	\vspace{-24pt}
	\label{fig:featurevis}
\end{figure}

%Fig.~\ref{fig:featurevis} shows that AAN can highlight adaptively the effective information from complex feature maps and effectively weakened interference factors caused by occlusion. In the end, the expression ability of feature maps is significantly improved.

\Remark By virtue of cross-ANN, the cross-interdependencies between $\mathbf{R}_{\rm A}$ and $\mathbf{R}_{\rm c}$ are aggregated. Figure~\ref{fig:featurevis} clearly shows that AAN adaptively highlights the effective information from complex feature maps and effectively weakens the interference factors caused by occlusion. Eventually, based on the improved feature map, the robustness and accuracy of SiamAPN++ are raised.

\subsubsection{Classification and regression network}

The classification and regression network apply a similar structure compared with the baseline, \textit{i.e.}, SiamAPN. The multi-classification structure is also adopted. The first branch aims to classify the anchors with a high intersection over union (IoU) score. The second branch concentrates on selecting the points on the feature map that fall in ground truth areas. The last branch considers the center distance between each point and ground truth center point inspired by \cite{9157720}. $w_{1}$, $w_{2}$, and $w_{3}$ are introduced for balancing different branches. For improving the performance of convergence, the loss function of regression is redesigned as:
\begin{equation}
\begin{aligned}
L_{\rm ious}=-(1-ious)*(\alpha-ious)*\mathrm{log}(ious)~,\\
%&ious=\dfrac{\mathbf{R}_{anc}\bigcap\mathbf{R}_{gt}}{\mathbf{R}_{anc}\bigcup\mathbf{R}_{gt}}~
\end{aligned}
\end{equation}
where $\alpha$ is a hyper-parameter, reflecting the tendentiousness to positive and negative samples while $ious$ represents the IoU score between the proposed anchors and ground truth bounding box. $\alpha$ is set in the range of 1 to 2. It aims to increase gradient at high loss position to accelerate the convergence process, and decrease gradient at low loss position to raise the convergence accuracy. In this way, the model obtains accurate and robust convergence results faster (epoch=25) compared with the baseline (epoch=37).

\section{Evaluations}\label{sec:EXPERIMENT}

In this section, SiamAPN++ is comprehensively evaluated on two well-known authoritative UAV tracking benchmarks, \textit{i.e.}, UAV20L~\cite{Mueller2016ECCV} and UAV123@10fps~\cite{Mueller2016ECCV}. Other 14 SOTA trackers are also included in the evaluation, \textit{i.e.},
SiamAPN (baseline)~\cite{fu2020siamese}, SiamRPN++~\cite{8954116}, DaSiamRPN~\cite{zhu2018distractor}, SiamFC++~\cite{xu2020siamfc++}, SiamFC~\cite{bertinettofully}, UDT~\cite{Wang_2019_Unsupervised}, UDT+~\cite{Wang_2019_Unsupervised}, TADT~\cite{Xin2019CVPR}, DSiam~\cite{guo2017learning}, CoKCF~\cite{zhang2017robust}, CF2~\cite{Ma-ICCV-2015}, CFNet~\cite{8100014}, ECO~\cite{8100216}, and KCC~\cite{wang2018kernel}. 

\Remark To better compare the performance of different tracking strategies, all Siamese-based trackers adopt the same backbone, \textit{i.e.}, AlexNet~\cite{krizhevsky2012imagenet} pre-trained on ImageNet~\cite{russakovsky2015imagenet}.

\subsection{Implementation details}
\label{subsec:EvaCri}
In the training process, the last three layers of the backbone are fine-tuned with a learning rate of 5$\times$$10^{-4}$ at first. The whole tracker are trained on the images extracted from COCO~\cite{lin2014microsoft}, ImageNet VID~\cite{russakovsky2015imagenet}, GOT-10K~\cite{huang2019got} and Youtube-BB~\cite{real2017youtube}. The stochastic gradient
descent (SGD) with a minibatch of $220$ pairs is applied. Besides, the momentum is set to $0.9$ and weight decay is set to $10^{-4}$. Following the baseline, the input size of the template image and search image are set to $3\times 127 \times 127$ pixels and $3\times 287 \times 287$ pixels. The tracking code is available at \url{https://github.com/vision4robotics/SiamAPN}.

The training process of the proposed method SiamAPN++ is implemented in Python using Pytorch on a PC with an Intel i9-9920X CPU, a 32GB RAM, and two NVIDIA TITAN RTX GPUs. For testing the feasibility and performance of SiamAPN++ on UAV tracking, an NVIDIA Jetson AGX Xavier is adopted as the real-world tests platform. Real-world tests validate the accuracy and robustness of SiamAPN++ with a speed of around 35 frames per second (FPS) without TensorRT acceleration.

% network is trained with SGD with a minibatch of $124$ pairs. The momentum of  is used.
%
%
%SiamAPN uses the AlexNet as the backbone with the parameters of the first two convolution layers frozen and only fine-tune the last three convolution blocks. There are a total of 50 epochs. For the first 10 epochs, the parameters of the feature extraction network are frozen to train the APN, feature fusion network, and multi-classification\&regression network. For the last 40 epoch, the whole network is end-to-end trained with learning rate decayed from $0.005$ to $0.0005$ in log space. Meanwhile,  During the process of training, except $\lambda_{cls1}$ is set to 1.2, other parameters, \textit{i.e.}, $\lambda_{cls2}$, $\lambda_{cls3}$, $\lambda_{loc1}$, and $\lambda_{loc2}$, are set to 1. As for the input size of the template patch and search patch, they are set to $127 \times 127$ pixels and $287 \times 287$ pixels, receptively. The tracking code is available at \url{https://github.com/vision4robotics/SiamAPN}. 
%
%SiamAPN++ is trained extracted Images from COCO~\cite{lin2014microsoft}, ImageNet VID~\cite{russakovsky2015imagenet}, GOT-10K~\cite{huang2019got} and Youtube-BB~\cite{real2017youtube} are extracted to train our tracker. The proposed method is implemented in Python using Pytorch on a PC with an Intel i9-9920X CPU, a 32GB RAM, and an Nvidia TITAN RTX GPU.

\begin{figure}[t]
	\centering
	\includegraphics[width=0.49\linewidth]{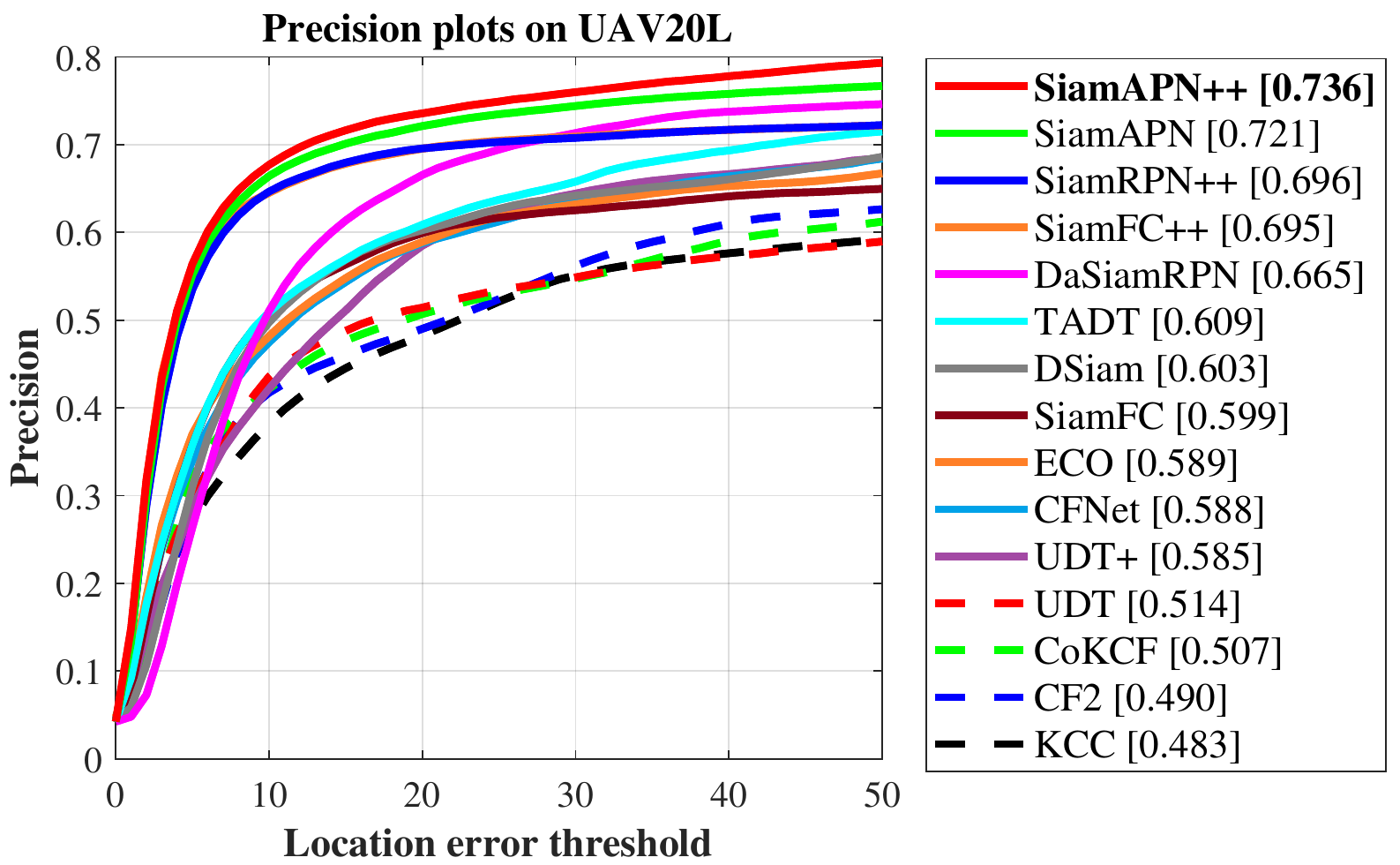}	
	\includegraphics[width=0.49\linewidth]{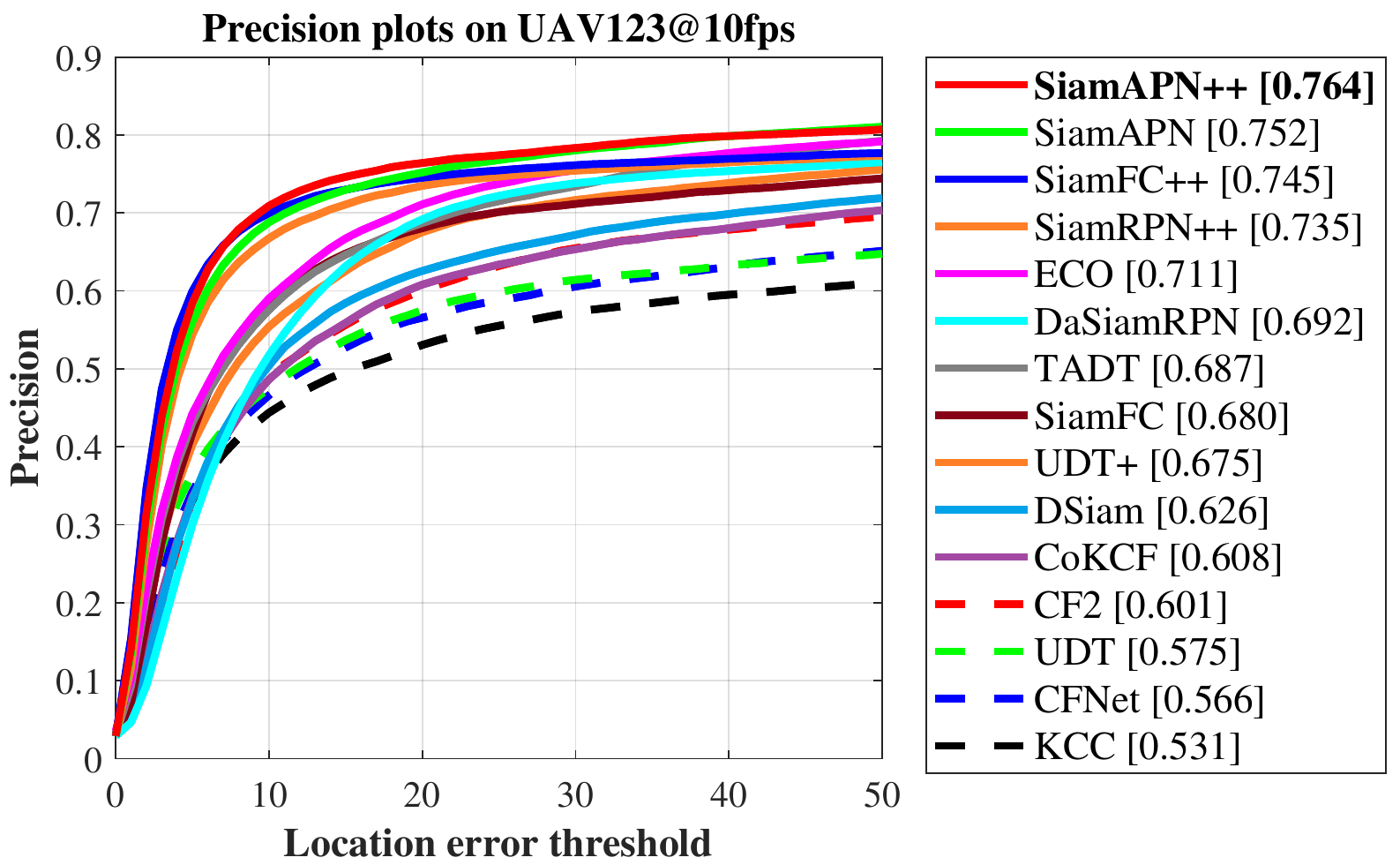}
	\\
	\vspace{-4pt}
	\subfloat[Results on UAV20L]
	{\includegraphics[width=0.5\linewidth]{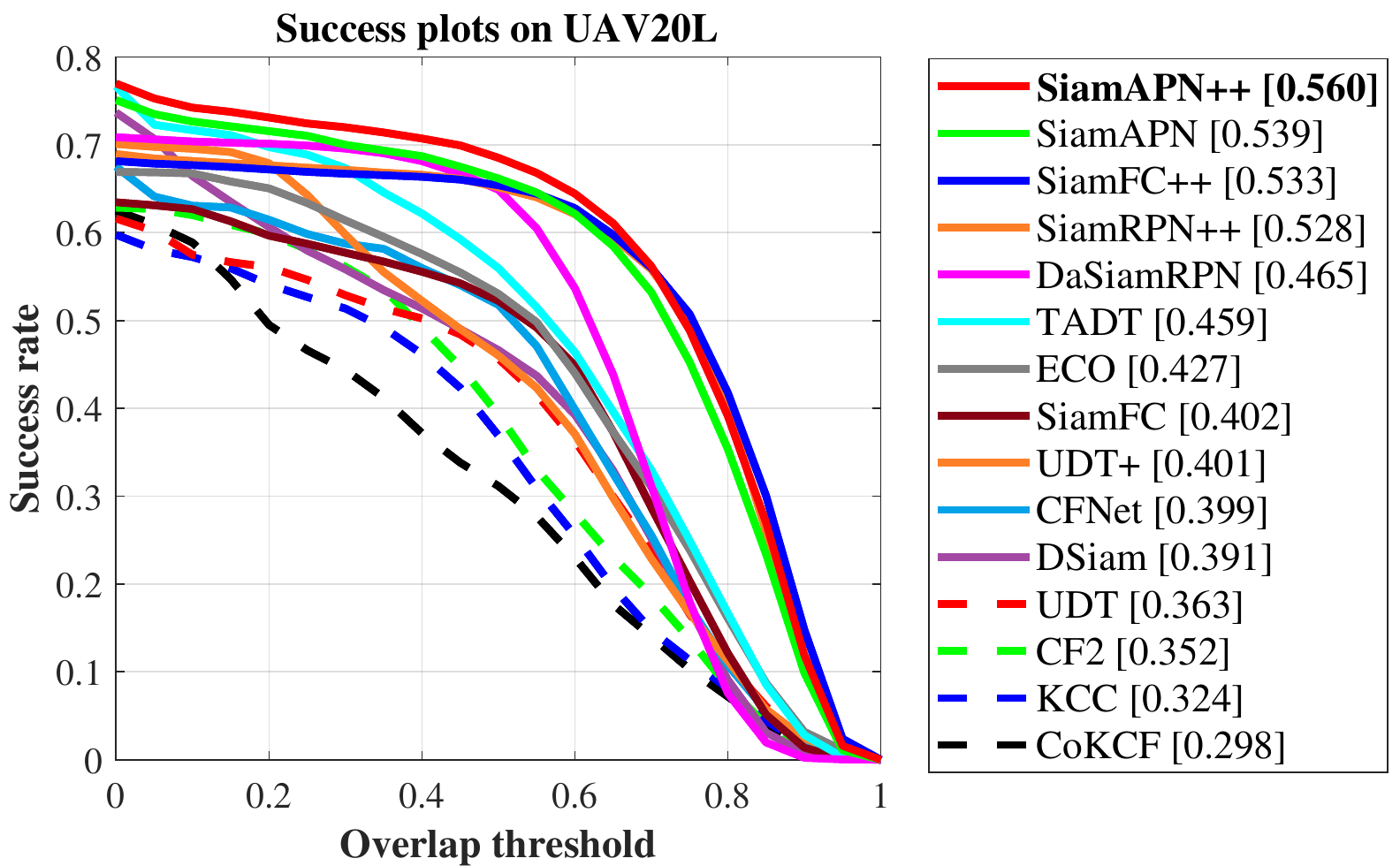}\label{fig:a}}
	\subfloat[Results on UAV123@10fps]
	{\includegraphics[width=0.5\linewidth]{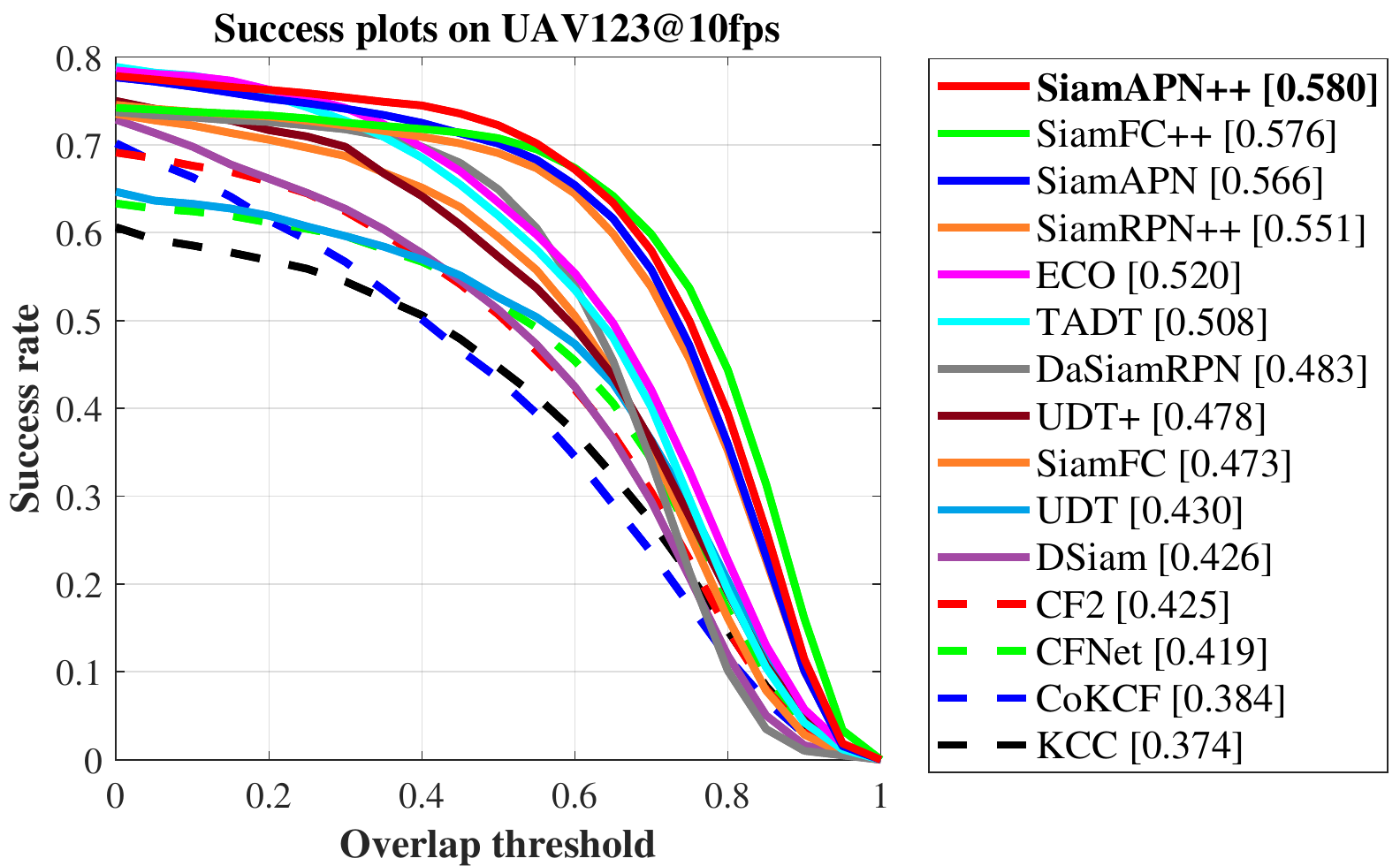}\label{fig:b}}
	
	\caption{Overall performance of all trackers on (a) UAV20L, (b) UAV123@10fps. The overall results illustrate that SiamAPN++ achieves superior performance against other SOTA trackers.}
	\vspace{-3pt}
	\label{fig:overperform}
\end{figure}
\begin{table}[t]
	
	\centering
	\caption{Average attribute-based evaluation of the SiamAPN++ and other 14 SOTA trackers on two benchmarks. The best three performances are respectively highlighted with \NoOne{\textbf{red}}, \NoTwo{\textbf{green}}, and \NoThree{\textbf{blue}} color.}
	\renewcommand\tabcolsep{5pt}
	\resizebox{0.98\linewidth}{!}{
	\begin{tabular}{lrrrrrrrrrr}
		\toprule[1.4pt]
		& \multicolumn{2}{c}{CM} & \multicolumn{2}{c}{FM} & \multicolumn{2}{c}{FOC} & \multicolumn{2}{c}{POC} & \multicolumn{2}{c}{SV} \\
		\midrule
		Trackers & \multicolumn{1}{l}{Prec.} & \multicolumn{1}{l}{Succ.} & \multicolumn{1}{l}{Prec.} & \multicolumn{1}{l}{Succ.} & \multicolumn{1}{l}{Prec.} & \multicolumn{1}{l}{Succ.} & \multicolumn{1}{l}{Prec.} & \multicolumn{1}{l}{Succ.} & \multicolumn{1}{l}{Prec.} & \multicolumn{1}{l}{Succ.} \\
		\midrule
		ECO   & 0.630 & 0.457 & 0.542 & 0.344 & 0.457 & 0.254 & 0.598 & 0.427 & 0.621 & 0.453 \\
		SiamFC & 0.626 & 0.432 & 0.529 & 0.319 & 0.487 & 0.264 & 0.577 & 0.385 & 0.614 & 0.416 \\
		UDT   & 0.512 & 0.375 & 0.470 & 0.303 & 0.413 & 0.226 & 0.493 & 0.350 & 0.513 & 0.375 \\
		UDT+  & 0.598 & 0.424 & 0.521 & 0.322 & 0.469 & 0.259 & 0.581 & 0.403 & 0.599 & 0.418 \\
		CF2   & 0.519 & 0.374 & 0.384 & 0.246 & 0.431 & 0.233 & 0.499 & 0.346 & 0.507 & 0.364 \\
		CFNet & 0.533 & 0.382 & 0.435 & 0.244 & 0.407 & 0.214 & 0.527 & 0.365 & 0.542 & 0.386 \\
		CoKCF & 0.525 & 0.329 & 0.414 & 0.210 & 0.418 & 0.205 & 0.512 & 0.309 & 0.520 & 0.311 \\
		DSiam & 0.613 & 0.404 & 0.536 & 0.309 & \NoTwo{\textbf{0.547}} & \NoThree{\textbf{0.301}} & 0.579 & 0.371 & 0.585 & 0.387 \\
		TADT  & 0.638 & 0.474 & 0.597 & 0.386 & 0.486 & 0.281 & 0.612 & 0.444 & 0.618 & 0.461 \\
		KCC   & 0.458 & 0.320 & 0.348 & 0.206 & 0.339 & 0.172 & 0.457 & 0.300 & 0.467 & 0.320 \\
		DaSiamRPN & 0.667 & 0.470 & 0.571 & 0.365 & 0.505 & 0.274 & 0.625 & 0.427 & 0.656 & 0.459 \\
		SiamFC++ & \NoThree{\textbf{0.709}} & \NoTwo{\textbf{0.544}} & 0.622 & \NoThree{\textbf{0.454}} & 0.469 & 0.292 & \NoThree{\textbf{0.669}} & \NoTwo{\textbf{0.497}} & \NoThree{\textbf{0.700}} & \NoTwo{\textbf{0.540}} \\
		SiamRPN++ & 0.706 & 0.530 & \NoThree{\textbf{0.624}} & 0.442 & 0.475 & 0.284 & 0.655 & 0.480 & 0.691 & 0.521 \\
		SiamAPN & \NoTwo{\textbf{0.721}} & \NoThree{\textbf{0.537}} & \NoTwo{\textbf{0.710}} & \NoTwo{\textbf{0.493}} & \NoThree{\textbf{0.520}} & \NoTwo{\textbf{0.308}} & \NoTwo{\textbf{0.680}} & \NoThree{\textbf{0.493}} & \NoTwo{\textbf{0.715}} & \NoThree{\textbf{0.537}} \\
		\midrule
		\textbf{SiamAPN++} & \NoOne{\textbf{0.742}} & \NoOne{\textbf{0.564}} & \NoOne{\textbf{0.711}} & \NoOne{\textbf{0.507}} & \NoOne{\textbf{0.577}} & \NoOne{\textbf{0.358}} & \NoOne{\textbf{0.702}} & \NoOne{\textbf{0.519}} & \NoOne{\textbf{0.729}} & \NoOne{\textbf{0.554}} \\
		\bottomrule[1.4pt]
	\end{tabular}}
	\label{tab:addlabel}%
	\vspace{-10pt}
\end{table}%

\begin{table}[t]
	\footnotesize
	\setlength{\tabcolsep}{1mm}
	\centering
	\caption{Precision comparison of our tracker with different components on UAV20L. Note that ARC represents the aspect ratio change and OC includes full occlusion as well as partial occlusion.}
	\centering
	{
		\begin{tabular}{lcccccc }
			\hline
			\hline

			{\textbf{Structure}}&{\textbf{Overall}}&SV&ARC&CM&OC\\ 
			\midrule
			Baseline & 0.721 &0.707&0.652&0.707&0.617\\
			Baseline+APN-DF & 0.727&0.715&0.662&0.713&0.649\\
			Baseline+APN-DF+self-AAN & 0.730&0.716&0.663&0.716&0.647\\
			Baseline+APN-DF+AAN & \textbf{0.736}& \textbf{0.722}& \textbf{0.670}& \textbf{0.722}& \textbf{0.651}\\
			\hline
			\hline
		\end{tabular}%
	}
	\label{tab:parameter}%
\end{table}%
%\begin{table*}[!h]
%	\footnotesize
%	%\setlength{\tabcolsep}{2.5mm}
%	\centering
%	
%	\caption{Average precision as well as AUC score of 15 trackers on two well known benchmarks. \textcolor[rgb]{ 1,  0,  0}{\textbf{Red}}, \textcolor[rgb]{ 0,  1,  0}{\textbf{green}}, and \textcolor[rgb]{ 0,  0,  1}{\textbf{Blue}} fonts indicate the first and second best results respectively.}
%	\centering
%	\setlength{\tabcolsep}{0.6mm}
%	{
%		\begin{tabular}{l c c c c ccccccccccc}
%			\hline
%			\hline
%			{\textbf{Trackers}}&{\textbf{SiamAPN++}}&SiamAPN&{SiamFC++}&{SiamRPN++}&{DaSiamRPN}&{TADT}&SiamFC&UDT+&CFNet&DSiam&ECO&UDT&CF2&KCC&CoKCF\\  
%			\hline
%			\textbf{Precision (\%)} & \textbf{\textcolor[rgb]{1  0  0}{75.0}} & \textbf{\textcolor[rgb]{0  1  0}{74.2}} & \textbf{\textcolor[rgb]{0  0  1}{72.0}} & {71.6}&{67.9}&64.8 & {63.4} &63.0 & {57.7}&{61.5}&65.0 & {54.5} & 54.6 & {50.7}&{55.8}
%			\\ 
%			\textbf{AUC (\%)} & \textbf{\textcolor[rgb]{1  0  0}{57.0}} & \textbf{\textcolor[rgb]{0  1  0}{55.9}} &\textbf{\textcolor[rgb]{0  0  1}{55.5}} & {54.0}&{47.4} & 48.4 & 43.8&44.0&40.9&40.9&47.4&39.7&38.9&34.9&34.1 \\
%			\hline
%			\hline
%		\end{tabular}%
%	}
%	\label{tab:fps}%
%\end{table*}%
\begin{figure}[h]
	\centering
	\includegraphics[width=1\linewidth]{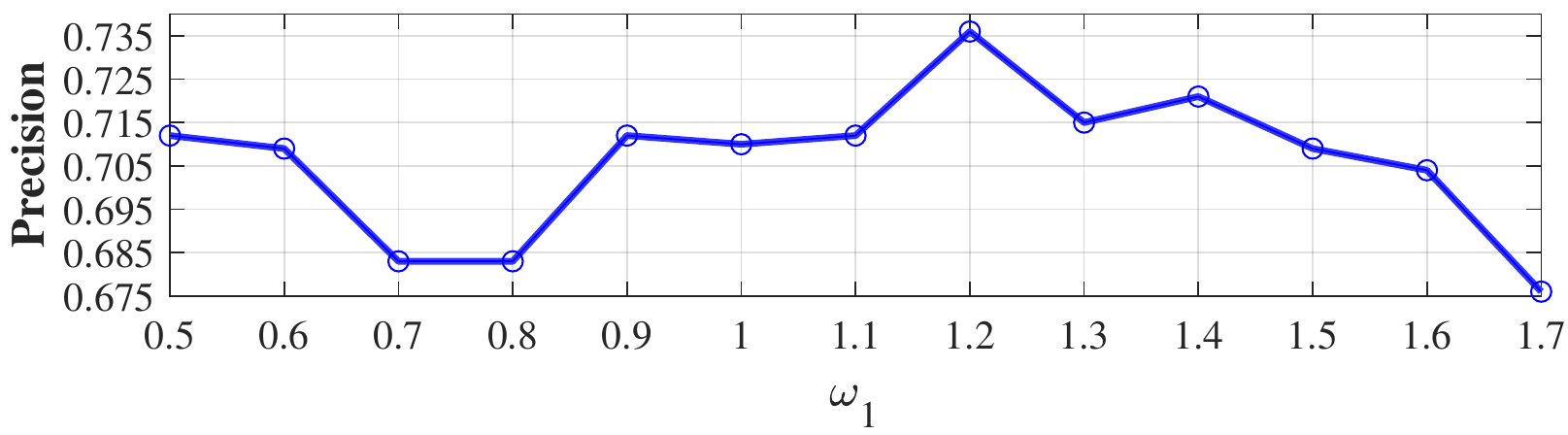}
	\includegraphics[width=1\linewidth]{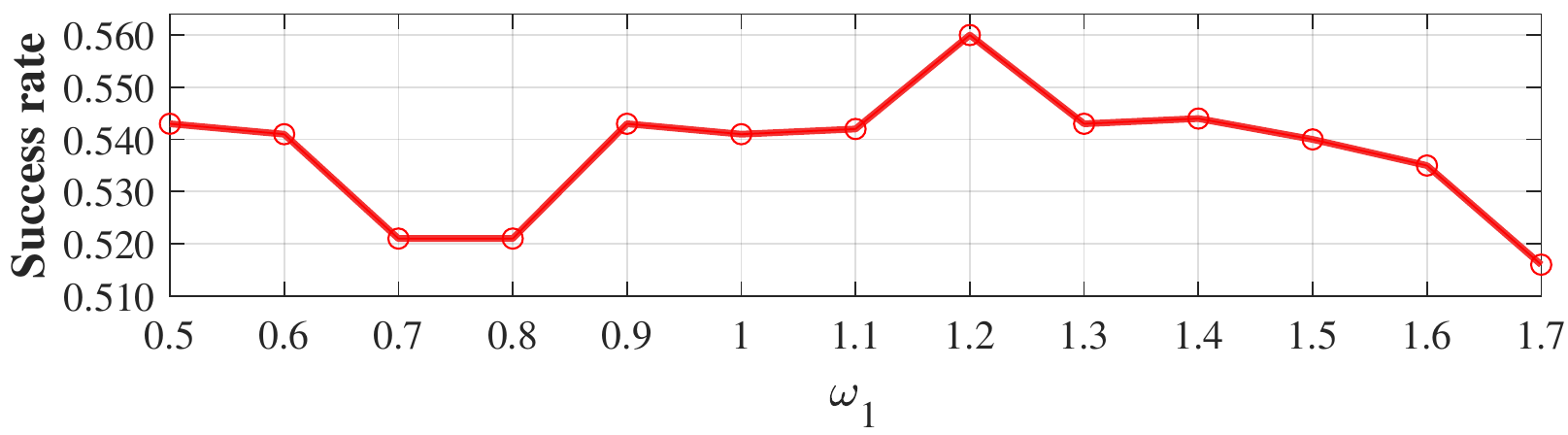}
	\vspace{-14pt}
	\caption{Key parameter analysis of $w_{1}$ on UAV20L. When the $w_{1}=1.2$, SiamAPN++ achieves the best overall performance, which is adopted in all experiments}
	\label{fig:parameters}
	\vspace{-10pt}
\end{figure}

\subsection{Evaluation metrics}
The one-pass evaluation (OPE) metrics~\cite{6619156} are adopted, \textit{i.e.}, precision and success rate. Specifically, the success rate is measured by the IoU score. Besides, the success plot reflects the percentage of the frames whose IoU score is beyond a pre-defined threshold and the area under the curve (AUC)
of success plot is used to rank all the trackers. The
precision plot shows the percentage of frames whose center location error (CLE) between the estimated bounding box and ground
truth is smaller than thresholds. Note that the score at 20 pixels is utilized for ranking.

%fast-motion, partical occlusion, full occlusion, and low resolution from UAV123@10fps~\cite{Mueller2016ECCV}, UAV20L~\cite{Mueller2016ECCV}, and VisDrone2018-SOT-test~\cite{wen2018visdrone}

\subsection{Evaluation on UAV benchmarks}
\subsubsection{Overall performance}
The proposed method achieves an impressive improvement compared with other SOTA trackers on two well-known benchmarks.

\textbf{UAV20L:}
UAV20L~\cite{Mueller2016ECCV} contains 20 long-term sequences whose maximum sequence contains 5527 frames with an average of 2934 frames per sequence. Therefore, UAV20L is adopted for evaluating long-term tracking performance. As illustrated in Fig.~\ref{fig:a}, SiamAPN++ outperforms other trackers with an improvement of \textbf{2.0\%} on precision and \textbf{4.0\%} on AUC score, compared with the second-best tracker.

\begin{figure}[b]
	\centering
	\vspace{-10pt}
	\includegraphics[width=1\linewidth]{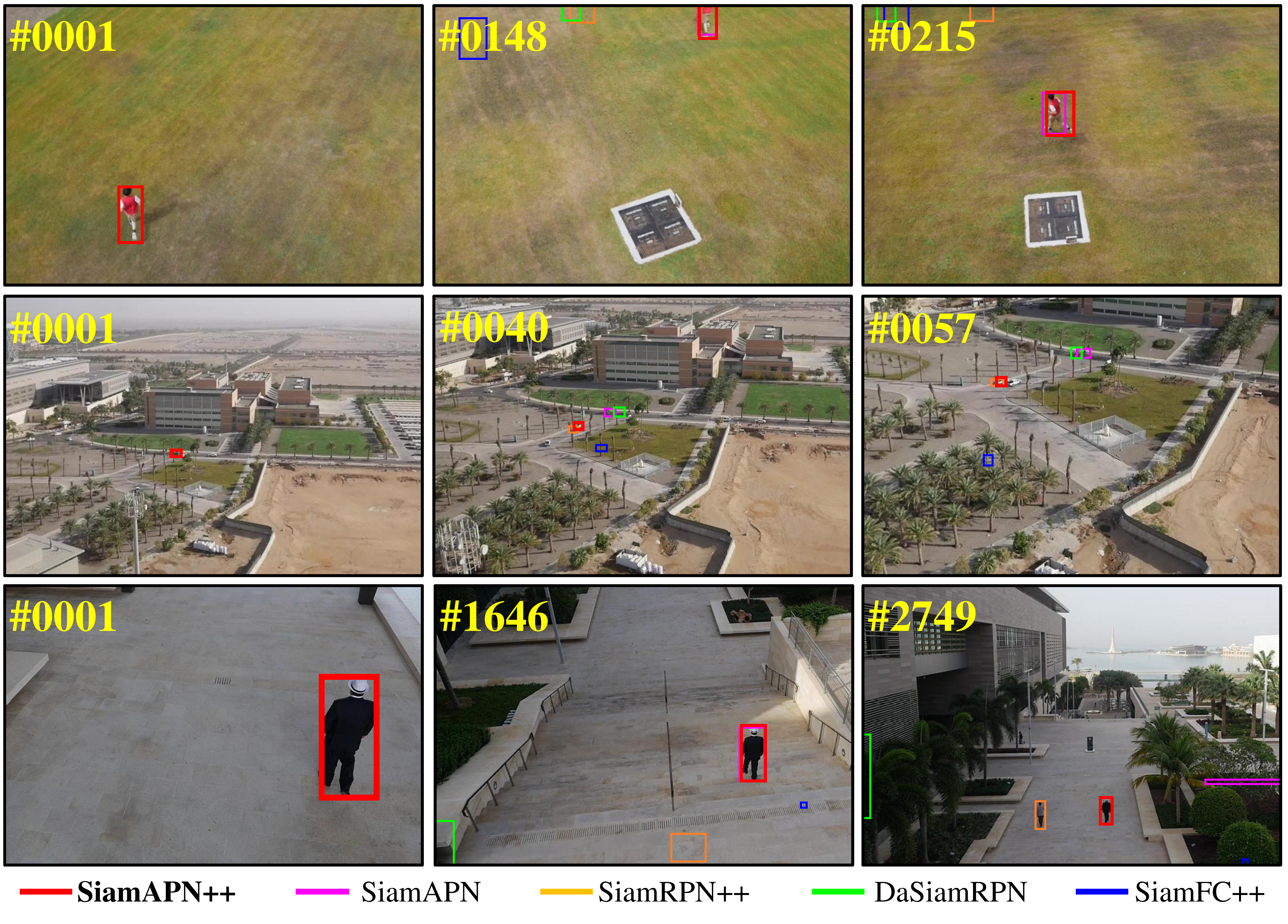}
    \vspace{-20pt}
	\caption{Screenshots of \textit{person7\_2}, \textit{car11} from UAV123@10fps, and \textit{person19} from UAV20L. The tracking videos can be found here: \url{https://youtu.be/okS289p3pCQ}.}

	\label{fig:5}
\end{figure}
\textbf{UAV123@10fps:}
UAV123@10fps~\cite{Mueller2016ECCV} contains 123 sequences with a frame rate of 10 FPS. Since the frame interval becomes larger than 30 FPS, the movement and variation of the object become more drastic, bringing difficulties to the tracking task. Therefore, UAV123@10fps~\cite{Mueller2016ECCV} is chosen to comprehensively evaluate the robustness of the tracker under severe variation. Overall performance shown in Fig.~\ref{fig:b} proves the superior robustness and accuracy of SiamAPN++ in precision (0.764) and AUC score (0.580).

%\begin{table}[t]
%	\footnotesize
%	\setlength{\tabcolsep}{1mm}
%	\centering
%	\caption{Precision comparison of our tracker with different structure on UAV20L. APN-HF represents the novel APN based on hybrid feature. SAN and CAN represent the respectively. Circle, Rectangle, Ellipse represent circle labels,
%		rectangle labels, ellipse labels, respectively.}
%	\centering
%	%\setlength{\tabcolsep}{0.8mm}
%	{
%		\begin{tabular}{l|cccccc }
%			\toprule[2pt]
%			
%			{\textbf{Structure}}&{\textbf{Overall}}&SV&ARC&CM&POC&FOC\\ 
%			\midrule
%			Baseline & 0.721 &0.707&0.652&0.707&0.698&0.536\\
%			Baseline+APN-HF & 0.727&0.715&0.662&0.713&0.704&\textbf{0.594}\\
%		    Baseline+APN-HF+SAN & 0.73&0.716&0.663&0.716&0.707&0.587\\
%			Baseline+APN-HF+SAN+CAN & \textbf{0.736}& \textbf{0.722}& \textbf{0.670}& \textbf{0.722}& \textbf{0.718}& 0.584\\
%
%
%			\bottomrule[2pt]
%		\end{tabular}%
%	}
%	\label{tab:parameter}%
%\end{table}%

\begin{figure}[t]
	\centering	
	\includegraphics[width=1\linewidth]{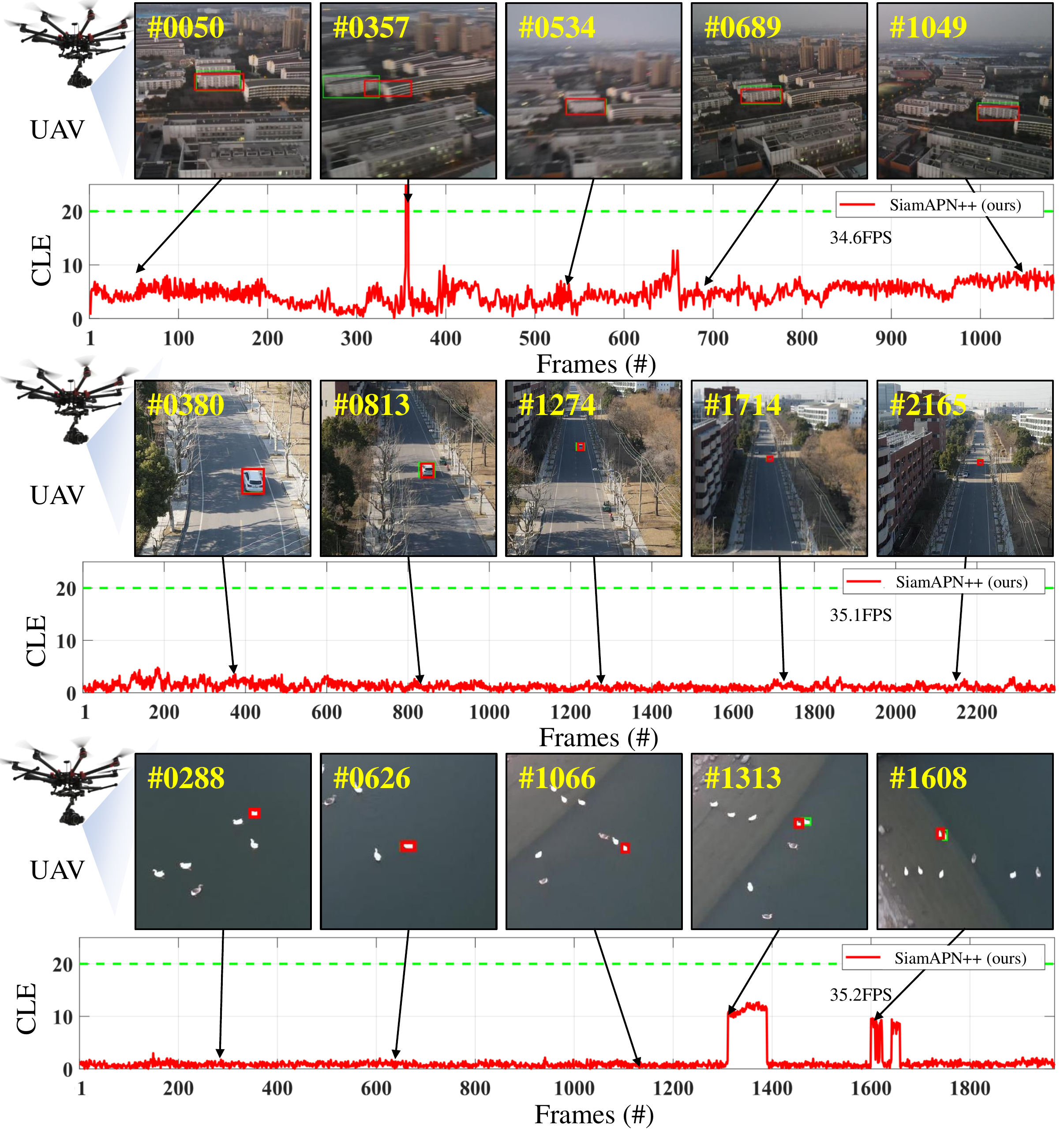}
	%	\subfloat[Partial occlusion fast motion, scale variation]
	%	{
	%		\includegraphics[width=0.32\linewidth]{images/real-world/1/car_eveningpre.pdf}
	%		\label{fig:a1}
	%	}
	%	\subfloat[Similar objects, severe camaera motion, partial occlusion ]
	%{
	%	%\includegraphics[width=0.32\linewidth]{images/real-world/1/building2_2pre.pdf}
	%	\includegraphics[width=1\linewidth]{images/real-world/s1.pdf}
	%	\label{fig:a1}
	%}\\
	%	\subfloat[Camera motion, low illumination, view point change]
	%	{
	%		\includegraphics[width=1\linewidth]{images/real-world/s1.pdf}
	%		%\includegraphics[width=0.32\linewidth]{images/real-world/1/bridge_darkpre.pdf}
	%		\label{fig:a2}
	%	}\\
	
	%	\subfloat[Camera motion, low resolution, viewpoint change]
	%{
	%	\includegraphics[width=0.32\linewidth]{images/real-world/1/car_night3pre.pdf}
	%	\label{fig:a4}
	%}
	%	\subfloat[low resolution, scale variation, aspect ratio change]
	%{
	%	\includegraphics[width=1\linewidth]{images/real-world/s1.pdf}
	%	%\includegraphics[width=0.32\linewidth]{images/real-world/1/car1pre.pdf}
	%	\label{fig:a3}
	%}
	
	%	\subfloat[similar object, low resolution, viewpoint change]
	%{
	%	\includegraphics[width=0.32\linewidth]{images/real-world/1/duck2_3pre.pdf}
	%	\label{fig:a6}
	%}

	\caption{Real-world tests in terms of CLE onboard the embedded platform. The tracking objects from the top to bottom are building, car, and duck. Besides, the tracking results and ground truth are marked with \NoOne{red} and \NoTwo{green} boxes. The CLE score below the \NoTwo{green} dotted line is considered as the effective tracking result. } 
	\vspace{-20pt}
	\label{fig:v4r}
\end{figure}

\subsubsection{Attribute-based performance}
\label{subsec:attr}
To analyze the robustness of SiamAPN++ under various challenges, the average attribute-based evaluation results on two UAV benchmarks are shown in TABLE~\ref{tab:addlabel}. Five most common attributes in UAV tracking challenges are analyzed, \textit{i.e.}, camera motion (CM), fast motion (FM), full occlusion (FOC), partial occlusion (POC), and scale variation (SV). Attributing to the AAN and dual feature structure, SiamAPN++ achieves impressive performance in the CM scenarios with a \textbf{5.0\%} promotion on AUC score. Besides, our tracker surpasses the baseline in terms of FM. Meantime, in the FOC conditions, SiamAPN++ exceeds the second-best tracker with a huge improvement of \textbf{5.5\%} in precision and \textbf{16.2\%} in AUC score.

\subsubsection{Ablation study}
To demonstrate the effectiveness of the APN-DF and AAN, the precision of SiamAPN++ with different components and the baseline SiamAPN on UAV20L is listed in TABLE~\ref{tab:parameter}. With the internal relationship introduced by the APN-DF, the tracker has surpassed the baseline. Besides, it indeed promotes performance when tracking objects with various scales. Furthermore, attributing to the AAN, the self-interdependencies from the single feature map and the cross-interdependencies are aggregated, further improving the accuracy of SiamAPN++.

\subsubsection{Key parameter analysis}
Since the first classification branch, \textit{i.e.}, $w_{1}$, directly reflects the classification tendentiousness to anchors, $w_{1}$ has important influence in tracking performance. Obviously, too large $w_{1}$ will influence the effectiveness of other branches, while too small $w_{1}$ will impede the promotion introduced by anchor information on classification. Therefore, to evaluate the influence of $w_{1}$, $w_{1}$ is set to different values for further research. It is set from 0.5 and increases in a small step of 0.1. As presented in Fig. \ref{fig:parameters}, the AUC and precision of SiamAPN++ achieve best performance when $w_{1}=1.2$. Please note that $w_{1}$ is set to 1.2 during the evaluation above and the real-world tests.

\subsubsection{Qualitative Evaluation}
Some qualitative comparisons are shown in Fig. \ref{fig:5}. The main challenges of these sequences include SV, ARC, FM, POC, CM, and out-of-view. By the combination of APN-DF and AAN, SiamAPN++ achieves superior tracking performance eventually.

\section{Real-World Tests}\label{sec:real}
To verify the feasibility of the proposed tracker, real-world tests are conducted onboard the embedded platform. Four real-world tests are presented in Fig.~\ref{fig:v4r} and Fig.~\ref{fig:1} includes car, building, and duck. The main challenges in the first test illustrated in Fig.~\ref{fig:v4r} are severe CM, POC, and similar objects. Although there are some errors due to the blurring caused by severe CM, SiamAPN++ can determine the location of the object again and achieve impressive long-term tracking performance. By aggregating self- and cross-interdependencies of feature maps, SiamAPN++ can handle the low resolution (LR), SV, ARC challenges effectively, thereby tracking the car accurately and robustly in the second scene. Meantime, benefiting from the location information of anchors, the robustness of SiamAPN++ is also improved. Based on it, the AAN can explore the effective interdependencies between feature maps, thereby achieving impressive tracking performance in the third test. Additionally, the test illustrated in Fig.~\ref{fig:1} mainly focuses on the occluded object, validating the effectiveness of the combination of AAN and APN-DF when facing severe occlusion. During the real-world tests, our tracker maintains robust and accurate performance with an average speed of 34.9 FPS. In a word, the real-world tests strongly prove the impressive performance of SiamAPN++ under various challenges with a promising speed in UAV tracking scenarios.

%%%%%%%%%%%%%%%%%%%%%%%%%%%%%%%%%%%%%%%%%%%%%%%%%%%%%%%%%%%%%%%%
%%%%%%%%%%%%%%%%%%%%% Section 5: CONCLUSIONS %%%%%%%%%%%%%%%%%%%
%%%%%%%%%%%%%%%%%%%%%%%%%%%%%%%%%%%%%%%%%%%%%%%%%%%%%%%%%%%%%%%%

\section{Conclusion}\label{sec:CONCLUSIONS}
In this work, a novel attentional Siamese-based tracker is introduced for fulfilling the performance and feasibility requirement of real-time UAV tracking. The self-AAN is proposed for aggregating the self-interdependencies of the single feature. Besides, to aggregate the cross-interdependencies between two different feature maps, we also propose the cross-AAN. In addition, the new dual feature structure also integrates different feature maps effectively. Exhaustive evaluations validate the effectiveness of our method. Meantime, real-world tests strongly verify the practicability of our tracker. Consequently, we believe that our work can boost the development of UAV tracking-related applications.

%Since the anchors can adapt to the movement and variation of objects, the location and size information of anchors are quite important for object location.
%%%%%%%%%%%%%%%%%%%%%%%%%%%%%%%%%%%%%%%%%%%%%%%%%%%%%%%%%%%%%%%%%%%%%%%%%%%%%%%%

%%%%%%%%%%%%%%%%%%%%%%%%%%%%%%%%%%%%%%%%%%%%%%%%%%%%%%%%%%%%%%%%%%%%%%%%%%%%%%%%
\vspace{-5pt}
\section*{Acknowledgment}
This work is supported by the National Natural Science Foundation of China (No. 61806148) and the Natural Science Foundation of Shanghai (No. 20ZR1460100).

%%%%%%%%%%%%%%%%%%%%%%%%%%%%%%%%%%%%%%%%%%%%%%%%%%%%%%%%%%%%%%%%%%%%%%%%%%%%%%%%

\bibliographystyle{IEEEtran}  %这是你要使用的格式,比如要投IEEE,就写IEEEtran
\bibliography{IEEEabrv,ref}%这个是加载你的bib,你可以理解从文献数据库中加载要引用的文献

\begin{thebibliography}{10}
\providecommand{\url}[1]{#1}
\csname url@rmstyle\endcsname
\providecommand{\newblock}{\relax}
\providecommand{\bibinfo}[2]{#2}
\providecommand\BIBentrySTDinterwordspacing{\spaceskip=0pt\relax}
\providecommand\BIBentryALTinterwordstretchfactor{4}
\providecommand\BIBentryALTinterwordspacing{\spaceskip=\fontdimen2\font plus
\BIBentryALTinterwordstretchfactor\fontdimen3\font minus
  \fontdimen4\font\relax}
\providecommand\BIBforeignlanguage[2]{{%
\expandafter\ifx\csname l@#1\endcsname\relax
\typeout{** WARNING: IEEEtran.bst: No hyphenation pattern has been}%
\typeout{** loaded for the language `#1'. Using the pattern for}%
\typeout{** the default language instead.}%
\else
\language=\csname l@#1\endcsname
\fi
#2}}

\bibitem{8968163}
R.~{Bonatti}, C.~{Ho}, W.~{Wang}, S.~{Choudhury}, and S.~{Scherer}, ``{Towards
  a Robust Aerial Cinematography Platform: Localizing and Tracking Moving
  Targets in Unstructured Environments},'' in \emph{Proceedings of the IEEE/RSJ
  International Conference on Intelligent Robots and Systems (IROS)}, 2019, pp.
  229--236.

\bibitem{8967602}
G.~J. {Laguna} and S.~{Bhattacharya}, ``{Path planning with Incremental Roadmap
  Update for Visibility-based Target Tracking},'' in \emph{Proceedings of the
  IEEE/RSJ International Conference on Intelligent Robots and Systems (IROS)},
  2019, pp. 1159--1164.

\bibitem{9457090}
J.~Ye, C.~Fu, F.~Lin, F.~Ding, S.~An, and G.~Lu, ``{Multi-Regularized
  Correlation Filter for UAV Tracking and Self-Localization},'' \emph{IEEE
  Transactions on Industrial Electronics}, pp. 1--10, 2021.

\bibitem{fu2020correlation}
C.~Fu, B.~Li, F.~Ding, F.~Lin, and G.~Lu, ``{Correlation Filter for UAV-Based
  Aerial Tracking: A Review and Experimental Evaluation},'' \emph{IEEE
  Geoscience and Remote Sensing Magazine}, pp. 1--28, 2020.

\bibitem{huang2019learning}
Z.~Huang, C.~Fu, Y.~Li, F.~Lin, and P.~Lu, ``{Learning Aberrance Repressed
  Correlation Filters for Real-time UAV Tracking},'' in \emph{Proceedings of
  the IEEE International Conference on Computer Vision (ICCV)}, 2019, pp.
  2891--2900.

\bibitem{Li_2020_CVPR}
Y.~{Li}, C.~{Fu}, F.~{Ding}, Z.~{Huang}, and G.~{Lu}, ``{AutoTrack: Towards
  High-Performance Visual Tracking for UAV With Automatic Spatio-Temporal
  Regularization},'' in \emph{Proceedings of the IEEE Conference on Computer
  Vision and Pattern Recognition (CVPR)}, 2020, pp. 11\,920--11\,929.

\bibitem{bertinettofully}
L.~Bertinetto, J.~Valmadre, J.~F. Henriques, A.~Vedaldi, and P.~H. Torr,
  ``{Fully-Convolutional Siamese Networks for Object Tracking},'' in
  \emph{Proceedings of the European Conference on Computer Vision (ECCV)},
  2016, pp. 850--865.

\bibitem{Li_2018_CVPR}
B.~{Li}, J.~{Yan}, W.~{Wu}, Z.~{Zhu}, and X.~{Hu}, ``{High Performance Visual
  Tracking with Siamese Region Proposal Network},'' in \emph{{Proceedings of
  the IEEE Conference on Computer Vision and Pattern Recognition (CVPR)}},
  2018, pp. 8971--8980.

\bibitem{8954116}
B.~{Li}, W.~{Wu}, Q.~{Wang}, F.~{Zhang}, J.~{Xing}, and J.~{Yan}, ``{SiamRPN++:
  Evolution of Siamese Visual Tracking With Very Deep Networks},'' in
  \emph{Proceedings of the IEEE Conference on Computer Vision and Pattern
  Recognition (CVPR)}, 2019, pp. 4277--4286.

\bibitem{zhu2018distractor}
Z.~Zhu, Q.~Wang, B.~Li, W.~Wu, J.~Yan, and W.~Hu, ``{Distractor-Aware Siamese
  Networks for Visual Object Tracking},'' in \emph{Proceedings of the European
  Conference on Computer Vision (ECCV)}, 2018, pp. 101--117.

\bibitem{9156275}
Y.~{Yu}, Y.~{Xiong}, W.~{Huang}, and M.~R. {Scott}, ``{Deformable Siamese
  Attention Networks for Visual Object Tracking},'' in \emph{Proceedings of the
  IEEE Conference on Computer Vision and Pattern Recognition (CVPR)}, 2020, pp.
  6727--6736.

\bibitem{xu2020siamfc++}
Y.~Xu, Z.~Wang, Z.~Li, Y.~Yuan, and G.~Yu, ``{SiamFC++: Towards Robust and
  Accurate Visual Tracking with Target Estimation Guidelines},'' in
  \emph{Proceedings of the AAAI Conference on Artificial Intelligence (AAAI)},
  2020, pp. 12\,549--12\,556.

\bibitem{9157720}
D.~{Guo}, J.~{Wang}, Y.~{Cui}, Z.~{Wang}, and S.~{Chen}, ``{SiamCAR: Siamese
  Fully Convolutional Classification and Regression for Visual Tracking},'' in
  \emph{Proceedings of the IEEE Conference on Computer Vision and Pattern
  Recognition (CVPR)}, 2020, pp. 6268--6276.

\bibitem{cao2021hift}
Z.~{Cao}, C.~{Fu}, J.~{Ye}, B.~{Li}, and Y.~{Li}, ``{HiFT: Hierarchical Feature
  Transformer for Aerial Tracking},'' in \emph{Proceedings of the IEEE
  International Conference on Computer Vision (ICCV)}, 2021, pp. 1--10.

\bibitem{fu2020siamese}
C.~Fu, Z.~Cao, Y.~Li, J.~Ye, and C.~Feng, ``{Siamese Anchor Proposal Network
  for High-Speed Aerial Tracking},'' in \emph{Proceedings of the IEEE
  International Conference on Robotics and Automation (ICRA)}, 2021, pp. 1--7.

\bibitem{8578608}
Q.~{Wang}, Z.~{Teng}, J.~{Xing}, J.~{Gao}, W.~{Hu}, and S.~{Maybank},
  ``{Learning Attentions: Residual Attentional Siamese Network for High
  Performance Online Visual Tracking},'' in \emph{Proceedings of the IEEE
  Conference on Computer Vision and Pattern Recognition (CVPR)}, 2018, pp.
  4854--4863.

\bibitem{5539960}
D.~S. {Bolme}, J.~R. {Beveridge}, B.~A. {Draper}, and Y.~M. {Lui}, ``{Visual
  Object Tracking Using Adaptive Correlation Filters},'' in \emph{Proceedings
  of the IEEE Conference on Computer Vision and Pattern Recognition (CVPR)},
  2010, pp. 2544--2550.

\bibitem{Fu_2020_TGRS}
C.~{Fu}, J.~{Ye}, J.~{Xu}, Y.~{He}, and F.~{Lin}, ``{Disruptor-Aware
  Interval-Based Response Inconsistency for Correlation Filters in Real-Time
  Aerial Tracking},'' \emph{IEEE Transactions on Geoscience and Remote
  Sensing}, pp. 1--13, 2020.

\bibitem{tao2016sint}
R.~{Tao}, E.~{Gavves}, and A.~W.~M. {Smeulders}, ``{Siamese Instance Search for
  Tracking},'' in \emph{Proceedings of the IEEE Conference on Computer Vision
  and Pattern Recognition (CVPR)}, 2016, pp. 1420--1429.

\bibitem{8579033}
B.~{Li}, J.~{Yan}, W.~{Wu}, Z.~{Zhu}, and X.~{Hu}, ``{High Performance Visual
  Tracking with Siamese Region Proposal Network},'' in \emph{Proceedings of the
  IEEE Conference on Computer Vision and Pattern Recognition (CVPR)}, 2018, pp.
  8971--8980.

\bibitem{9477413}
C.~Fu, Z.~Cao, Y.~Li, J.~Ye, and C.~Feng, ``{Onboard Real-Time Aerial Tracking
  With Efficient Siamese Anchor Proposal Network},'' \emph{IEEE Transactions on
  Geoscience and Remote Sensing}, pp. 1--13, 2021.

\bibitem{8578911}
X.~{Wang}, R.~{Girshick}, A.~{Gupta}, and K.~{He}, ``{Non-local Neural
  Networks},'' in \emph{Proceedings of the IEEE Conference on Computer Vision
  and Pattern Recognition (CVPR)}, 2018, pp. 7794--7803.

\bibitem{8953974}
J.~{Fu}, J.~{Liu}, H.~{Tian}, Y.~{Li}, Y.~{Bao}, Z.~{Fang}, and H.~{Lu},
  ``{Dual Attention Network for Scene Segmentation},'' in \emph{Proceedings of
  the IEEE Conference on Computer Vision and Pattern Recognition (CVPR)}, 2019,
  pp. 3141--3149.

\bibitem{9156697}
Q.~{Wang}, B.~{Wu}, P.~{Zhu}, P.~{Li}, W.~{Zuo}, and Q.~{Hu}, ``{ECA-Net:
  Efficient Channel Attention for Deep Convolutional Neural Networks},'' in
  \emph{Proceedings of the IEEE Conference on Computer Vision and Pattern
  Recognition (CVPR)}, 2020, pp. 11\,531--11\,539.

\bibitem{krizhevsky2012imagenet}
A.~Krizhevsky, I.~Sutskever, and G.~E. Hinton, ``{Imagenet Classification with
  Deep Convolutional Neural Networks},'' in \emph{Advances in neural
  information processing systems (NIPS)}, 2012, pp. 1097--1105.

\bibitem{Mueller2016ECCV}
M.~Mueller, N.~Smith, and B.~Ghanem, ``{A Benchmark and Simulator for UAV
  Tracking},'' in \emph{Proceedings of the European Conference on Computer
  Vision (ECCV)}, 2016, pp. 445--461.

\bibitem{Wang_2019_Unsupervised}
N.~{Wang}, Y.~{Song}, C.~{Ma}, W.~{Zhou}, W.~{Liu}, and H.~{Li},
  ``{Unsupervised Deep Tracking},'' in \emph{Proceedings of the IEEE Conference
  on Computer Vision and Pattern Recognition (CVPR)}, 2019, pp. 1308--1317.

\bibitem{Xin2019CVPR}
X.~{Li}, C.~{Ma}, B.~{Wu}, Z.~{He}, and M.~{Yang}, ``{Target-Aware Deep
  Tracking},'' in \emph{Proceedings of the IEEE Conference on Computer Vision
  and Pattern Recognition (CVPR)}, 2019, pp. 1369--1378.

\bibitem{guo2017learning}
Q.~{Guo}, W.~{Feng}, C.~{Zhou}, R.~{Huang}, L.~{Wan}, and S.~{Wang},
  ``{Learning Dynamic Siamese Network for Visual Object Tracking},'' in
  \emph{Proceedings of the IEEE International Conference on Computer Vision
  (ICCV)}, 2017, pp. 1781--1789.

\bibitem{zhang2017robust}
L.~Zhang and P.~N. Suganthan, ``{Robust Visual Tracking via Co-Trained
  Kernelized Correlation Filters},'' \emph{Pattern Recognition}, vol.~69, pp.
  82--93, 2017.

\bibitem{Ma-ICCV-2015}
C.~{Ma}, J.~{Huang}, X.~{Yang}, and M.~{Yang}, ``{Hierarchical Convolutional
  Features for Visual Tracking},'' in \emph{Proceedings of the IEEE
  International Conference on Computer Vision (ICCV)}, 2015, pp. 3074--3082.

\bibitem{8100014}
J.~{Valmadre}, L.~{Bertinetto}, J.~{Henriques}, A.~{Vedaldi}, and P.~H.~S.
  {Torr}, ``{End-to-End Representation Learning for Correlation Filter Based
  Tracking},'' in \emph{Proceedings of the IEEE Conference on Computer Vision
  and Pattern Recognition (CVPR)}, 2017, pp. 5000--5008.

\bibitem{8100216}
M.~{Danelljan}, G.~{Bhat}, F.~S. {Khan}, and M.~{Felsberg}, ``{ECO: Efficient
  Convolution Operators for Tracking},'' in \emph{Proceedings of IEEE
  Conference on Computer Vision and Pattern Recognition (CVPR)}, 2017, pp.
  6931--6939.

\bibitem{wang2018kernel}
C.~Wang, L.~Zhang, L.~Xie, and J.~Yuan, ``{Kernel Cross-Correlator},'' in
  \emph{Proceedings of the AAAI Conference on Artificial Intelligence (AAAI)},
  vol.~32, no.~1, 2018, pp. 4179--4186.

\bibitem{russakovsky2015imagenet}
O.~Russakovsky, J.~Deng, H.~Su, J.~Krause, S.~Satheesh, S.~Ma, Z.~Huang,
  A.~Karpathy, A.~Khosla, M.~Bernstein, \emph{et~al.}, ``{Imagenet Large Scale
  Visual Recognition Challenge},'' \emph{International Journal of Computer
  Vision}, vol. 115, no.~3, pp. 211--252, 2015.

\bibitem{lin2014microsoft}
T.-Y. Lin, M.~Maire, S.~Belongie, J.~Hays, P.~Perona, D.~Ramanan,
  P.~Doll{\'a}r, and C.~L. Zitnick, ``{Microsoft coco: Common objects in
  context},'' in \emph{Proceedings of the European conference on computer
  vision (ECCV)}, 2014, pp. 740--755.

\bibitem{huang2019got}
L.~{Huang}, X.~{Zhao}, and K.~{Huang}, ``{GOT-10k: A Large High-Diversity
  Benchmark for Generic Object Tracking in the Wild},'' \emph{IEEE Transactions
  on Pattern Analysis and Machine Intelligence}, pp. 1--17, 2019.

\bibitem{real2017youtube}
E.{Real}, J.{Shlens}, S.{Mazzocchi}, X.{Pan}, and V.{Vanhoucke},
  ``{YouTube-BoundingBoxes: A Large High-Precision Human-Annotated Data Set for
  Object Detection in Video},'' in \emph{Proceedings of the IEEE Conference on
  Computer Vision and Pattern Recognition (CVPR)}, 2017, pp. 7464--7473.

\bibitem{6619156}
Y.~{Wu}, J.~{Lim}, and M.~{Yang}, ``{Online Object Tracking: A Benchmark},'' in
  \emph{Proceedings of the IEEE Conference on Computer Vision and Pattern
  Recognition}, 2013, pp. 2411--2418.

\end{thebibliography}

\end{document}